\documentclass[
]{ceurart}

\begin{document}

\copyrightyear{2023}
\copyrightclause{Copyright for this paper by its authors.
  Use permitted under Creative Commons License Attribution 4.0
  International (CC BY 4.0).}

\conference{Forum for Information Retrieval Evaluation, December 15-18, 2023, India}

\title{Hate Speech and Offensive Content Detection in Indo-Aryan Languages: A Battle of LSTM and Transformers.}

\author[1]{Nikhil Narayan}[%
email=nikhilnarayan73@gmail.com,
]
\address[1]{Z-AGI Labs, India}

\author[1]{Mrutyunjay Biswal}[%
email=mrutyunjay.biswal.hmu@gmail.com,
]

\author[1]{Pramod Goyal}[%
email=goyalpramod1729@gmail.com,
]

\author[1]{Abhranta Panigrahi}[%
email=abhranta.panigrahi@gmail.com
]

\begin{abstract}
Social media platforms serve as accessible outlets for individuals to express their thoughts and experiences, resulting in an influx of user-generated data spanning all age groups. While these platforms enable free expression, they also present significant challenges, including the proliferation of hate speech and offensive content. Such objectionable language disrupts objective discourse and can lead to radicalization of debates, ultimately threatening democratic values. Consequently, social media platforms have taken steps to monitor and curb abusive behavior, necessitating automated methods for identifying suspicious posts. This paper contributes to Hate Speech and Offensive Content Identification in English and Indo-Aryan Languages (HASOC) 2023 shared tasks track for Hate Speech Detection in Low-Resource Languages. We, team Z-AGI Labs, conduct a comprehensive comparative analysis of hate speech classification across five distinct languages—Bengali, Assamese, Bodo, Sinhala, and Gujarati—within the context of the HASOC competition. Our study encompasses a wide range of pre-trained models, including Bert variants, XLM-R, and LSTM models, to assess their performance in identifying hate speech across these languages. Results reveal intriguing variations in model performance. Notably, Bert Base Multilingual Cased emerges as a strong performer across languages, achieving an F1 score of 0.67027 for Bengali and 0.70525 for Assamese. At the same time, it significantly outperforms other models with an impressive F1 score of 0.83009 for Bodo. In Sinhala, XLM-R stands out with an F1 score of 0.83493, whereas for Gujarati, a custom LSTM-based model outshined with an F1 score of 0.76601. This study offers valuable insights into the suitability of various pre-trained models for hate speech detection in multilingual settings. By considering the nuances of each, our research contributes to an informed model selection for building robust hate speech detection systems.
\end{abstract}


\maketitle

\section{Introduction}
In the era of expanding global connectivity through social media, platforms such as Facebook, X (Formerly Twitter), YouTube, and Instagram have grappled with a disturbing surge in hate speech perpetuated by individuals and organized groups. With the surge of Influencers and Content Creators, also commonly known as the Creator Economy\cite{enwiki:1159137601}, there has been an alarming rise in incidents concerning targeted attacks on individuals based on their opinions, appearance, and ethnicity\cite{thewire}. The consequences of such pervasive abusive language are far-reaching, often resulting in public humiliation and significant personal and professional consequences\cite{10.1001/jamapsychiatry.2019.2325, articleMentalhealth, articleCyberBullying} for victims. The escalating prevalence of online hate speech, often characterized by anonymity, scale, and overwhelming volumes that challenge human moderators, underscores the pressing need for social media platforms to strike a balance between safeguarding freedom of expression and fostering an environment of inclusiveness and respect.

The need for content moderation\cite{mitSloanSchool} by detecting Hate and Offensive engagement has pushed organizations and research groups to develop systems and solutions at scale. Significant work has been done to identify toxic, profane, and offensive comments. However, a majority of contributions focus predominantly on Resource-heavy languages such as English\cite{8637460, Vaidya_Mai_Ning_2020, Carta2019ASM, tran2020habertor, kamal2021hostility, hitkul2020trawling, founta2018large}. This brings constraints to hate and offensive speech content detection and moderation in low-resource languages. The lack of large-scale corpus and pre-trained models makes it extremely difficult to tackle Natural Language Understanding (NLU) downstream tasks.

In response to these challenges, Hate Speech and Offensive Content Identification in English and Indo-Aryan Languages (HASOC) presented four shared tasks as a part of its 5th edition, 2023. Out of these, Task 1 and Task 4\cite{ACM_Hasoc2023TaskOverview1&4} focus on detecting hate and offensive content in 5 low-resource languages as follows:- Bodo\cite{annihilate-hates-bodo}, Bengali\cite{annihilate-hates-bengali}, and Assamese\cite{annihilate-hates-assamese} in Task 4\cite{Hasoc2023TaskOverview4}, Gujarati and Sinhala in Task 1\cite{Hasoc2023TaskOverview1}. Each language has its corresponding dataset, evaluation metric, competition page, and leaderboard. The challenges present one problem statement, that is to classify a given content into one of the following classes:
\begin{itemize}
    \item \textbf{HOF} (Hate and/or Offensive): This content is hate speech, offensive, and/or profane.
    \item \textbf{NOT} (Non-Hate and/or Offensive): This content is not hate speech, offensive, and/or profane.
\end{itemize}

In this paper, we describe our approach to tackle the challenges. Here’s the summary of our contribution:
\begin{itemize}
    \item Adequate preprocessing techniques for datasets.
    \item Provide a strong LSTM with an attention head baseline model.
    \item Comparative analysis of pre-trained models in zero-shot and few-shot settings.
    \item Fine-tuning large multi-lingual models on the given datasets.
\end{itemize}

From here, the report continues in the following manner: In section 2, we highlight previous approaches as related work. In section 3, we give an overview of the dataset for each language and describe the challenge at hand. In section 4, we present our approach in detail, covering the nitty gritty of our experimental set-up, cross-validation strategy, models used, and intuition behind them. In section 5, we brief the results from the experiments section. Then, we conclude in section 6 with the final takeaways, our standings, and the scope of future work. The implementation details can be found in the following GitHub repository\footnote{https://github.com/The-Originalz/fire-hasoc-2023}.

\section{Related Work}
Detecting hate speech poses a formidable challenge in the realm of research, with existing literature encompassing diverse methodologies, such as dictionary-based approaches\cite{4618794}, the utilization of distributional semantics\cite{10.1145/2740908.2742760}, and the recent exploration of the efficacy of neural network architectures\cite{10.1145/3041021.3054223}. However, it is notable that a substantial portion of this research predominantly focuses on hate speech detection within the English language. Conversely, there is limited scholarly attention directed towards other foreign languages\cite{leite-etal-2020-toxic, saroj-pal-2020-indian, basile-etal-2019-semeval, ghosh-chowdhury-etal-2019-arhnet} and the intricacies of code-switched text\cite{ramitShawneyAAAI20, Kapoor_Kumar_Rajput_Shah_Kumaraguru_Zimmermann_2019}. Despite the significant impact of regional low-resource languages on online hate speech, this domain remains relatively uncharted, with recent investigations exploring the utility of transformers\cite{NIPS2017_3f5ee243} and author profiling through the application of graph neural networks\cite{ranasinghe2020wlvrit}.

Historically, numerous strategies have been employed to address the challenge of identifying hate speech. Kwok and Wang\cite{10.5555/2891460.2891697} experimented with a straightforward bag of words (BOW) methodology to detect hate speech, but these lightweight models yielded subpar results characterized by elevated false positive rates. Enhancing these models with various fundamental natural language processing (NLP) components, such as part-of-speech tags\cite{6406271} and N-gram graphs, contributed to improved performance. Lexical techniques employing TF-IDF in conjunction with Support Vector Machines (SVM) as a classification model surprisingly achieved commendable results\cite{Rajalakshmi2020DLRGHASOC2A}.

The advent of embedding words into distributed representations marked a pivotal shift, as researchers harnessed word embeddings like Glove\cite{pennington-etal-2014-glove} and FastText\cite{bojanowski-etal-2017-enriching} to project discrete text into a latent space, surpassing the performance of conventional BOW and lexical approaches.

Recurrent Neural Networks (RNNs) remained the go-to method for tackling various natural language challenges over an extended period. For instance, the winning approach in the 2020 HASOC competition for Hindi\cite{Raj2020NSITI} employed a one-layer Bidirectional Long Short-Term Memory (BiLSTM) model with FastText embeddings to discern hate speech. Likewise, the most accurate model for English\cite{mishra-2020-iiitdwd} adopted an LSTM architecture with Glove embeddings to represent textual inputs. Mohtaj et al.\cite{Mohtaj2020TUBAH} also embraced a character-based LSTM, aligning with this prevailing trend.

In recent times, self-attention-based transformer models\cite{NIPS2017_3f5ee243}, and their derivatives such as BERT\cite{DBLP:conf/naacl/DevlinCLT19}, derived from extensive corpus-trained encoders, have exhibited greater potential than traditional RNNs across a multitude of NLP tasks. BERT-like models have garnered substantial attention due to their remarkable transfer learning capabilities, outperforming alternative approaches consistently\cite{inbook19}.

Despite the substantial body of research on hate speech detection, experiments dedicated to low-resource languages remain relatively scarce. Notably, simple logistic regression using LASER embeddings demonstrated superior performance to BERT-based models\cite{aluru2020deep}, underscoring the necessity for more precise multilingual base language models. Consequently, we have witnessed the ascendancy of multilingual language models like XLM-Roberta\cite{conneau-etal-2020-unsupervised}. Following the trend, region-specific low-resource language models are developed. Some of the notable contributions are MuRIL\cite{khanuja2021muril}, SinBERT\cite{dhananjaya-etal-2022-bertifying} for Sinhala, BanglaBERT\cite{hasan-etal-2020-low}, Indic-BERT\cite{doddapaneni2023leaving}, and XLM-Indic\cite{moosa-etal-2023-transliteration} variants. Authors in \cite{ghosh-senapati-2022-hate} provide a detailed study on performance of mono-lingual and multi-lingual in the context of cross-lingual evaluation for hate speech identification. Previous editions of HASOC\cite{10.1145/3574318.3574326, 10.1145/3503162.3503176} have witnessed a significant effort towards improving performance in low-resource languages such as Hindi\cite{bhatia2021rule}, Marathi\cite{ranasinghe2022overview}, etc. In the ensuing sections, we will elucidate our approach, which leverages several multilingual models for hate speech identification, accompanied by an exhaustive comparative analysis against alternative methodologies.

\section{Dataset Description}
\subsection{Sinhala}
Task 1 consists of 2 sub-tasks,  one for Sinhala, and the other for Gujarati. Sinhala, one of the two official languages in Sri Lanka, is spoken by over 17 million people. This edition of HASOC brings the first-ever shared task for the aforementioned Indo-Aryan low-resource language. The train and test sets for Sinhala are based on the SOLD: Sinhala Offensive Language Detection dataset\cite{ranasinghe2022sold}. The SOLD consists of 10,000 manually annotated tweets divided into two classes: Offensive and Not offensive, both at the token level and the sentence level. However, the dataset provided for the task contains 7500 samples in the train set, each labeled as HOF or NOT. The test set contains 2500 samples.

\begin{figure}
  \centering
  \label{train-test-split}
  \includegraphics[width=\linewidth]{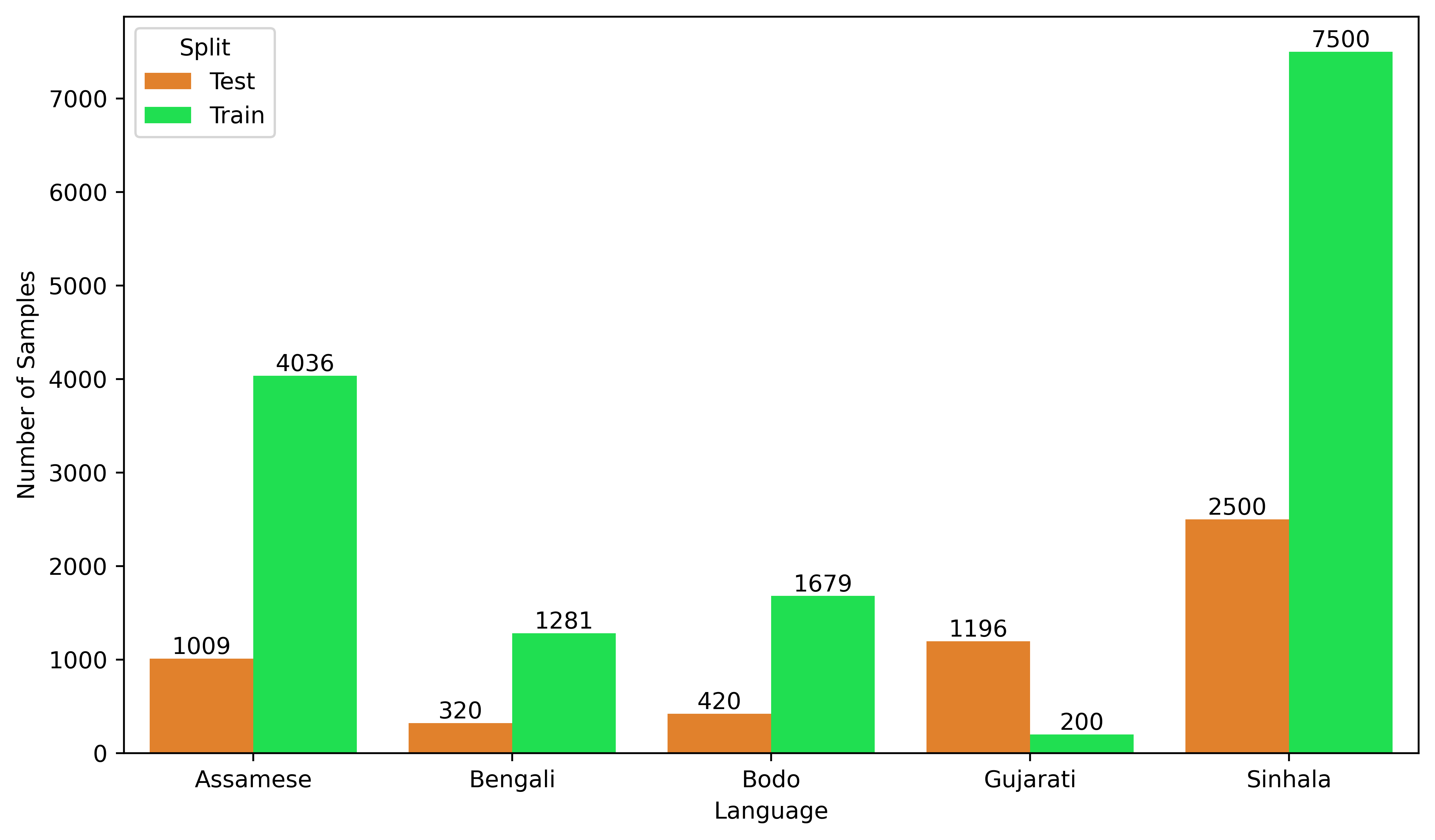}
  \caption{Train-Test Split for the respective Tasks}
\end{figure}
\subsection{Gujarati}
Gujarati is one of the 22 official languages of India with over 50M native speakers. The train set for this task contains 200 tweets whereas the test set contains 1196 tweets. This is also a coarse-grained binary classification problem but in a few-shot setting. The train data frame is made up of 5 columns, named as follows: tweet\_id, created\_at, text, user\_screen\_time, and label. The test set contains only tweet\_id and the text column. 

\subsection{Assamese, Bodo, and Bengali}
Task 4 consists of 3 Kaggle competitions, each corresponding to one of the following languages: Assamese, Bodo, and Bengali. The primary sources for data collection are X (formerly Twitter), Facebook, and YouTube Comments. Each train set contains tweet-label pairs, with only tweets in the test set to predict the targets.

\begin{figure}
  \centering
  \label{target-dist}
  \includegraphics[width=\linewidth]{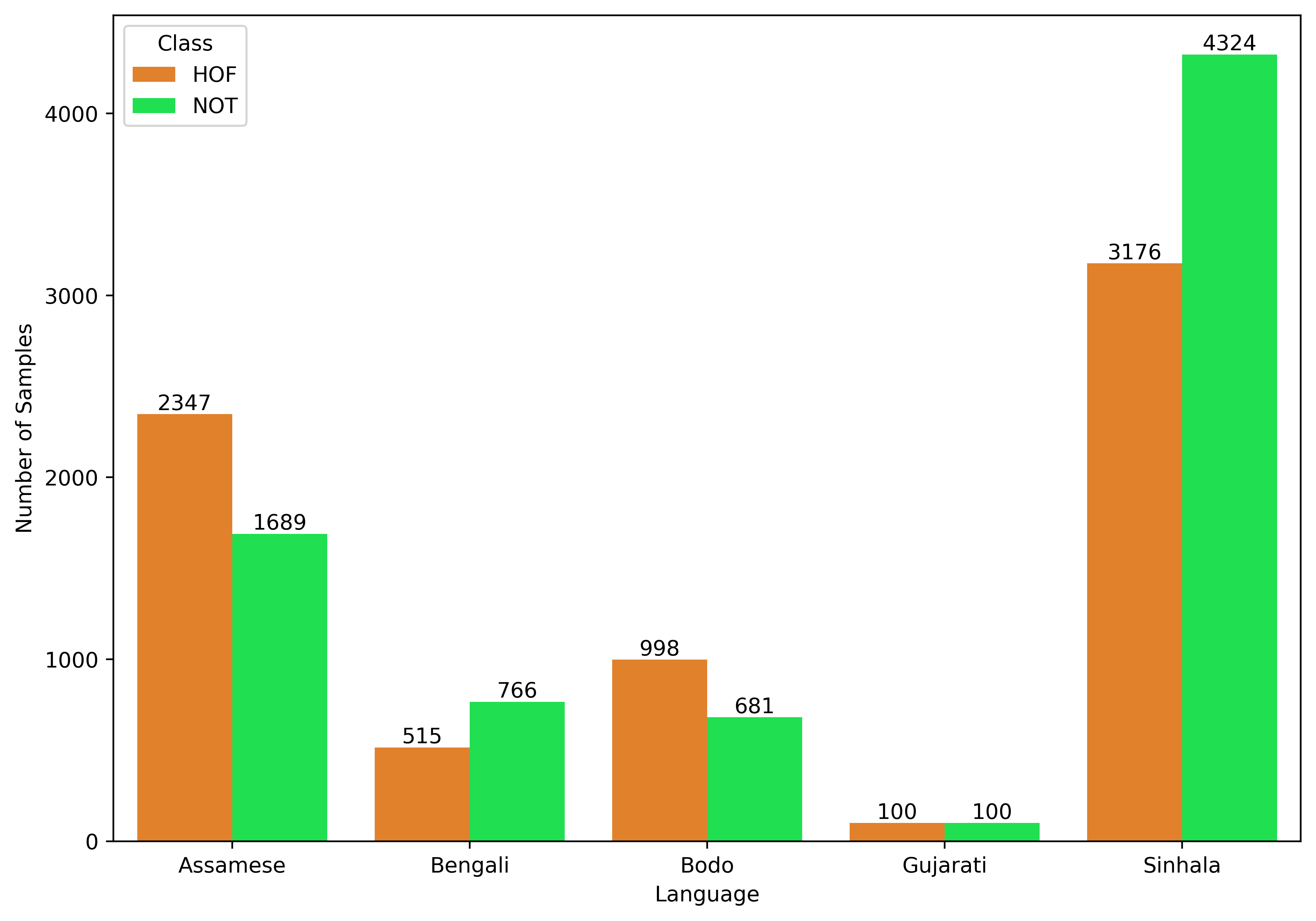}
  \caption{Target Distribution for the respective tasks.}
\end{figure}
\section{Experimental Set-up}

In this section, we discuss our approach and explain the experimental set-up details. We start with creating a validation strategy for each dataset. As all the datasets are fairly balanced (refer to figure \ref{target-dist}), we opt for K-Fold cross-validation with 5 folds. And while creating the splits, we set the random seed to 2023.

\subsection{Preprocessing}
The preprocessing step involves cleaning the contents for feature extraction. We notice that the Bodo tweets do not contain any emoji or repetitive punctuation marks that need attention. However, there are instances where some of the content contains English words in a code-mixed manner. Note that, such instances are seen across all the datasets. For the other two datasets in Task 4, though the majority of the content is cleaned, some of them contain emojis and a few repetitive punctuation characters. The repetition is removed, and the emojis are converted into their respective textual description using the emot\footnote{https://github.com/NeelShah18/emot} library.
The Task 1 datasets for Sinhala and Gujarati contain usernames as @USER and the usernames are made available in a separate column. Along with that, the datasets also contain code-mixed English words, repetitive punctuation characters, emojis, and hashtags. Note that, there are no URLs or hyperlinks associated with any of the content. The usernames are removed along with repetitive punctuation characters. The hashtags are further processed with Ekphrasis\footnote{https://github.com/cbaziotis/ekphrasis} tokenizer to segment them into meaningful tokens. The emoji2vec\cite{eisner-etal-2016-emoji2vec} embeddings are used on top of emojis, as it has shown competitive results in previous works.

\subsection{Modeling}
To start with, we create an LSTM-attention-based baseline with 2 bi-direction LSTM layers followed by an attention block. The attention head is further connected through 2 dense layers with sigmoid activation in the last layer. The model is trained with Adam optimizer and Binary Cross Entropy as the loss function. The hyperparameters involved such as the batch size, number of epochs, learning rate, vocab size, embedding dimension, and maximum length of the input sequence are varied and tuned on a case-to-case basis.

As the available datasets have less number of samples per language (refer to figure\ref{train-test-split}), we also leverage Transformer-based language models for the downstream task at hand. The available training data are used to fine-tune the encoder layers of the transformer-based models, leaving the embedding layers frozen. Note that to incorporate emojis semantics, we concatenate the embedding layers with the emoji2vec-generated embeddings. We experiment with various multi-lingual transformer-based models for fine-tuning such as Bert-Base-Multilingual (Cased and Uncased), DistilBert-Base-Multilingual-Cased, XLM-Roberta-Base, Muril-Base, and XLM-Indic-Base (UniScript\footnote{https://huggingface.co/ibraheemmoosa/xlmindic-base-uniscript} and Multi-Script\footnote{https://huggingface.co/ibraheemmoosa/xlmindic-base-multiscript}). Other than that, we also present the results from a couple of language-specific models such as Bangla-BERT and SinhalaBERTo\footnote{https://huggingface.co/keshan/SinhalaBERTo}.
\begin{table*}
\caption{Model aliases used a reference in the result graphs.}
\label{model-alias}
\begin{tabular}{cccl}
\toprule
Model Name & Alias Name \\
\midrule
LSTM Baseline & LSTM \\
Bert Base Multilingual Cased & mBert C \\
Bert Base Multilingual Uncased & mBert U \\
DistilBert Base Multilingual Cased & mDistil C \\
XLM Roberta Base & XLM-R \\
Muril Base Cased & MuRIL \\
XLM Indic Base Multiscript & XLM-I M \\
XLM Indic Base Uniscript & XLM-I U \\
\bottomrule
\end{tabular}
\end{table*}

Note that, the Huggingface implementation for the models\footnote{https://huggingface.co/models} is used via TFAutoModelForSequenceClassification with corresponding hyperparameters for each. All the training and inference are done using Kaggle runtime and MacBook Air M2 with 24GB unified memory.

For inference, we ensemble the models from each fold with equal weightage on the logits, then take a threshold of 0.5 to classify into HOF or NOT. The labels are mapped to numbers as follows: HOF is mapped to 0, and NOT is mapped to 1.

\section{Results}
\begin{figure}
  \label{res-viz}
  \centering
  \includegraphics[width=\linewidth]{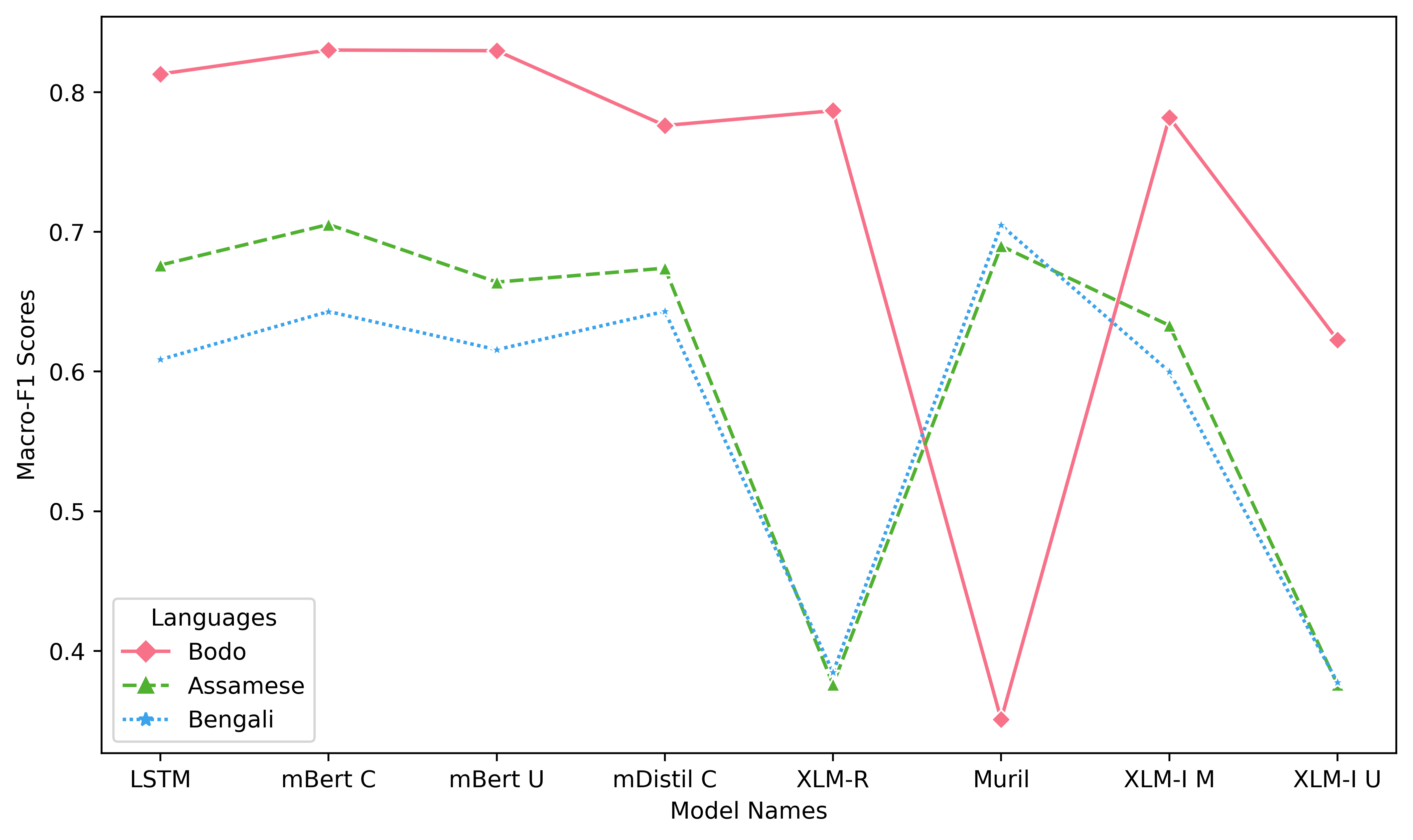}
  \caption{Leaderboard macro-F1 scores of different models for respective languages in Task 4.}
\end{figure}

The competitions for Task 4 are evaluated on macro-f1 metrics, whereas the Task 1 challenges are evaluated on macro-f1, Precision, and Recall.

\begin{table*}
\caption{Leaderboard macro-f1 scores in Task 4.}
\label{res-task4}
\begin{tabular}{cccl}
\toprule
Models & Bodo & Assamese & Bengali \\
\midrule
LSTM Baseline & 0.81291 & 0.67616 & 0.60856 \\
Bert Base Multilingual Cased & \textbf{0.83009} & \textbf{0.70525} & 0.64294 \\
Bert Base Multilingual Uncased & 0.82962 & 0.66398 & 0.61561 \\
DistilBert Base Multilingual Cased & 0.77606 & 0.67399 & 0.643 \\
XLM Roberta Base & 0.78668 & 0.376 & 0.38476 \\
Muril Base Cased & 0.35085 & 0.68985 & 0.70467 \\
XLM Indic Base Multiscript & 0.78186 & 0.633 & 0.59989 \\
XLM Indic Base Uniscript & 0.62257 & 0.376 & 0.37743 \\
csebuetnlp/banglabert & - & - & \textbf{0.75625} \\
\bottomrule
\end{tabular}
\end{table*}

It is evident from the results matrix (refer to table \ref{res-task1}) that, the LSTM baseline poses a strong competition in performance for all the languages. Even going a step further for Gujarati, the LSTM-based Model scores highest amongst XLM-Indic-Base-MultiScript and Bert-Base-Multilingual. This results in the highest F1-Score of 0.76601 and a Recall of 0.79704, with a marginal gain of 0.001 for F1 and 0.03 for Recall against the second-best performing model for the task. For Sinhala, XLM-Roberta seems to be the winner beating our LSTM Baseline and Sinhala Bert with a considerable margin. Due to time constraints and run submission limits, we experimented with a handful of BERT-based and Roberta-based models for fine-tuning along with the LSTM-with-Attention baseline Model.

For Task 4, we have varied candidate models for experimentation as mentioned in section 4 (refer to table \ref{model-alias}). Our LSTM baseline poses as one of the top performers for Bodo and Assamese by yielding the third-highest F1-Score, beating XLM-Roberta-based models. For Bengali, however, BanglaBert beat the rest of the models with a staggering 0.75625 f1-score. The best-performing model for Bodo and Assamese is Bert-Base-Multilingual-Cased with an f1-score of 0.83009, and 0.70525 respectively. Note that, all the scores mentioned above (refer to table\ref{res-task4}) are the performance on the hidden test set, and directly taken from the system-run report provided by the Organizers after a finalized leaderboard. A visual representation of comparative performance for Task 4 is shown in figure \ref{res-viz}.

\begin{table*}
\caption{Leaderboard macro-f1 scores in Task 1.}
\label{res-task1}
\begin{tabular}{ccl}
\toprule
Models & Gujarati & Sinhala \\
\midrule
LSTM Baseline & \textbf{0.7660} & 0.7530 \\
Bert Multilingual base Cased & 0.7656 & 0.8095 \\
XLM Roberta Base & - & \textbf{0.8349} \\
XLM Indic Base Multiscript & 0.7235 & - \\ 
\bottomrule
\end{tabular}
\end{table*}

The obtained results help us climb to 3rd/20 for Bengali, 5th/16 for Sinhala, 7th/17 for Gujarati, 8th/20 for Assamese, and 12th/19 for Bodo.

\section{Conclusion}
This work has been submitted to the CEUR-2023 Workshop Proceedings for Hate Speech and Offensive Content Identification in English and Indo-Aryan Languages (HASOC) track. In this report, we entail our approach to solving two tasks for the Track on classifying a given content into Hate and Offensive (HOF) or NOT. We experiment with various models starting from simple LSTM-based architecture to pretrained transformer-based multilingual models. Though the transformer shows sheer dominance in the majority of the tasks, our LSTM baseline emerges as a strong competitor with promising results, even resulting in the highest f1-score amongst our candidate models for Gujarati. We plan to further extend our work to other low-resource indic languages, and explore the possibilities of zero-shot, few-shot, and cross-lingual transfer learning scenarios. We also aim to develop a unified language model to incorporate all the languages such as NLLB\cite{nllbteam2022language} for similar downstream tasks to strengthen content moderation. 

\bibliography{references}

\end{document}